\documentclass[letterpaper, 10 pt, conference]{ieeeconf}  

\IEEEoverridecommandlockouts                              

\usepackage{multirow}
\usepackage{hyperref}
\usepackage{graphicx}
\usepackage{array}
\newcolumntype{C}[1]{>{\centering\arraybackslash}m{#1}}
\usepackage{hhline}
\usepackage{amsfonts}
\usepackage{amsmath}
\usepackage[caption=false]{subfig}
\usepackage{booktabs}
\usepackage{tikz}

\title{\LARGE \bf
A Spatio-Temporal Multilayer Perceptron for Gesture Recognition
}

\author{Adrian Holzbock$^{1\dag}$, Alexander Tsaregorodtsev$^{1\dag}$, Youssef Dawoud$^{1}$, Klaus Dietmayer$^{1}$ \\ and Vasileios Belagiannis$^{2*}$
\thanks{Part of the work was supported by INTUITIVER (7547.223-3/4/), funded by State Ministry of Baden-Württemberg for Sciences, Research and Arts and the State Ministry of Transport Baden-Württemberg, as well as, the LUKAS project (19A20004F), funded by German Federal Ministry for Economic Affairs and Energy (BMWi).}
\thanks{$^{1}$Institute of Measurement, Control and Microtechnology,
        Ulm University, Albert-Einstein-Allee 41, 89081 Ulm, Germany.}
\thanks{$^{2}$Department of Simulation and Graphics,
        Otto von Guericke University Magdeburg, Universitätsplatz 2, 39106 Magdeburg, Germany.}
\thanks{Email: firstname.lastname@uni-ulm.de, firstname.lastname@ovgu.de.}
\thanks{\dag\ denotes equal contribution.}
\thanks{*Most of this work was done while Vasileios Belagiannis was with Ulm University.}
}

\begin{document}

\newcommand\copyrighttextinitial{
    \scriptsize This work has been submitted to the IEEE for possible publication. Copyright may be transferred without notice, after which this version may no longer be accessible.}
    
\newcommand\copyrighttextfinal{
    \scriptsize\copyright\ 2022 IEEE. Personal use of this material is permitted. Permission from IEEE must be obtained for all other uses, in any current or future media, including reprinting/republishing this material for advertising or promotional purposes, creating new collective works, for resale or redistribution to servers or lists, or reuse of any copyrighted component of this work in other works. DOI: 10.1109/IV51971.2022.9827054.}
    
\newcommand\copyrightnotice{
    \begin{tikzpicture}[remember picture,overlay]
    \node[anchor=south,yshift=10pt] at (current page.south) {{\parbox{\dimexpr\textwidth-\fboxsep-\fboxrule\relax}{\copyrighttextfinal}}};
    \end{tikzpicture}
}

\maketitle
\copyrightnotice
\thispagestyle{empty}
\pagestyle{empty}

\begin{abstract}

Gesture recognition is essential for the interaction of autonomous vehicles with humans. While the current approaches focus on combining several modalities like image features, keypoints and bone vectors, we present neural network architecture that delivers state-of-the-art results only with body skeleton input data. We propose the spatio-temporal multilayer perceptron for gesture recognition in the context of autonomous vehicles. Given 3D body poses over time, we define temporal and spatial mixing operations to extract features in both domains. Additionally, the importance of each time step is re-weighted with Squeeze-and-Excitation layers. An extensive evaluation of the TCG and Drive\&Act datasets is provided to showcase the promising performance of our approach. Furthermore, we deploy our model to our autonomous vehicle to show its real-time capability and stable execution.

\end{abstract}


\section{INTRODUCTION}

Gestures are necessary for the interaction between autonomous vehicles and humans. For example, traffic control officers can request an autonomous vehicle to stop or turn with specific hand gestures. Similarly, the driver can control it from the inside of a car, e.g.~pointing to a desired parking spot. Because of its high importance, the problem of gesture recognition for autonomous vehicles is not new to the community~\cite{Chaman2018, bergasa2006real, liang2007real}. The current state-of-the-art for indoor~\cite{drive_and_act_2019_iccv} and outdoor~\cite{wiederer2020traffic} gesture recognition builds on deep neural networks. A popular approach is to extract the 2D or 3D body poses from images~\cite{bouazizi2021learning, bouazizi2021self, belagiannis2014holistic}, which is considered a more robust representation for gesture recognition~\cite{zimek2012survey, jhuang2013towards}. Then a neural network can learn from the skeleton-based data over time~\cite{wiederer2020traffic,yan2018spatial}. Also, convolutional neural networks are often used to directly recognize gestures from image data~\cite{carreira2017quo,qiu2017learning}. Alternatively, multiple streams of neural network models are considered for processing the spatial and temporal dimensions of the body pose, as well as the image data over time~\cite{martin2018body}. In this work, we show that only using the body pose representation over time is sufficient for the gesture recognition task in the context of autonomous vehicles.

\begin{figure}[!ht]
\vspace{2mm}
    \centering
    \includegraphics[width=0.48\textwidth]{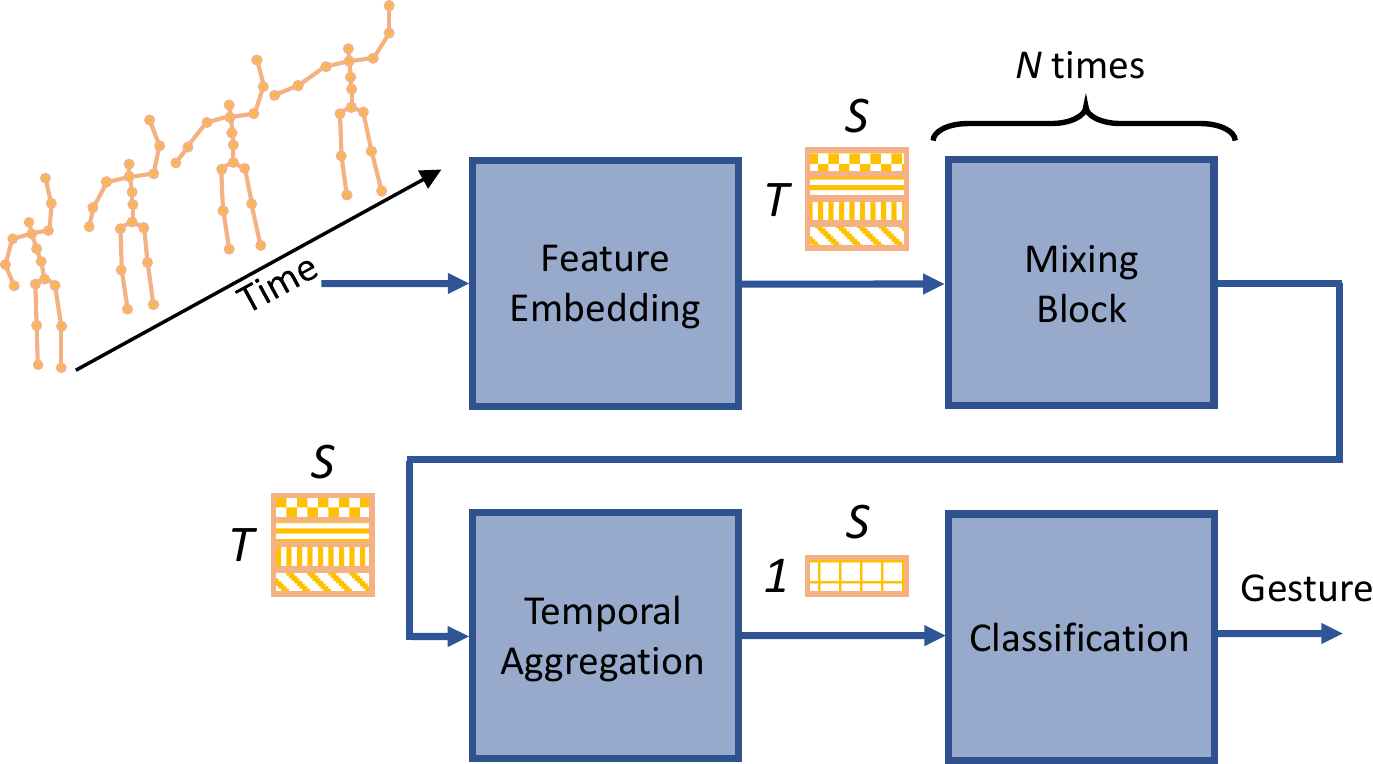}
    \caption{Overview of our proposed spatio-temporal MLP (st-MLP). Features are extraced from the time sequence of 3D human body poses and processed with the following $N$ mixing blocks. The squares represent the matrices where $T$ are the number of time steps and $S$ the dimension of the encoded skeleton features. The output of the mixing blocks is aggregated over time and is used for the gesture classification.}
    \label{fig:overview_stmlp}
\end{figure}

We present the spatio-temporal multilayer perceptron (st-MLP) for autonomous vehicle gesture recognition. A visual overview of our approach is given in Fig.~\ref{fig:overview_stmlp}. Unlike prior work, we process the spatial and temporal domain of the 3D body pose sequence simultaneously with a single neural network. To design our model, we derive our motivation from MLP-Mixers \cite{tolstikhin2021mlp}, which were designed for image classification. An MLP-Mixer performs feature mixing operations between the two image dimensions. Based on this observation, we define the temporal and spatial mixing operations to process 3D body poses over the time and space dimensions. Moreover, the advantage of the MLP-Mixers, instead of 3D convolutional neural networks, is the reduction of required computational power, as each dimension is mixed separately. First, our \textit{spatial-mixing} extracts features in the spatial domain. Next, our \textit{temporal-mixing} processes the data in the temporal domain. In addition, we introduce the Squeeze-and-Excitation (SE) block \cite{Hu_2018_CVPR} after each mixing operation to re-calibrate the feature representation and to shift attention. Our model learns a joint feature representation based on several blocks of interchangeable temporal and spatial mixing operations.

The design of our architecture aims at a low-latency deep neural network dedicated to automated driving gesture recognition. For that reason, we pick the TCG \cite{wiederer2020traffic} and Drive\&Act \cite{drive_and_act_2019_iccv} datasets for evaluation which contains typical gestures in the field of autonomous driving. In our experiments, we demonstrate state-of-the-art results when comparing with related work on both datasets. In addition, we compress our model and deploy it to our autonomous vehicle to show the real-time capabilities of our approach.

In summary, we propose an efficient approach for gesture recognition in automated driving, while we are the first to use an MLP-Mixer architecture for temporal data processing in the field of autonomous driving. Finally, we demonstrate real-time performance in our autonomous vehicle. Our code and pre-trained models are publicly available\footnote[1]{\url{https://github.com/holzbock/st_mlp}}.

\section{RELATED WORK}

\subsection{Gesture Recognition for Autonomous Driving}

The capability to understand gestures is a key enabler for automated driving applications where gestures involve body parts motion like hands, arms, head, and/or the entire body~\cite{mitra2007gesture}. Over the past few years, the gesture recognition problem in the vicinity of automated driving has been well-studied in the context of diverse human-vehicle interaction applications~\cite{rautaray2015vision,rasouli2019autonomous}. This includes interactions inside the vehicle~\cite{Sachara2017,pickering2007research,ohn2013vehicle}, e.g.~control of infotainment systems, as well as interactions outside the vehicle, e.g.~traffic control officers and pedestrians~\cite{Zengler2018,gupta2016conventionalized}. Further studies address other aspects of the problem, such as the lack of public datasets on traffic control gestures~\cite{wiederer2020traffic}, relying on robust motion capture sensors~\cite{geng2020using}, driver behavior prediction~\cite{martin2019drive}, and pedestrian intention prediction~\cite{8317766, 8370119, 8500425}. These works shed light on the importance of an accurate understanding of the human body language for automated driving. Furthermore, they make use of deep neural networks as their gateway towards achieving state-of-the-art results~\cite{nunez2018convolutional,lindgren2018learned,Molchanov2015}. Unlike prior work in neural networks, we present a model based on MLP-Mixers~\cite{tolstikhin2021mlp} using no recurrent structures or convolutional layers and instead only relying on linear layers and a transpose operation for skeleton-based gesture recognition. This enables our approach to processes 3D skeleton data in the spatial and temporal domain using only a single stream, similar to~\cite{yan2018spatial}. In contrast, an LSTM~\cite{hochreiter1997long} can only process the data in the temporal dimension. We compare the results of our approach with results of temporal models e.g.~Bi-LSTMs~\cite{zou2019deep} on the TCG~\cite{wiederer2020traffic} and Drive\&Act~\cite{drive_and_act_2019_iccv} datasets and show significant improvements.

In autonomous vehicles, real-time performance is also important for a safe and user-friendly experience, motivated by certain applications like driver behaviour prediction for handover control~\cite{deo2019looking} or pedestrian behaviour interpretation at crosswalks~\cite{deb2018investigating}. Although existing works often claim real-time capabilities with deep neural networks~\cite{Ehrnsperger2020,Choi2019}, the execution time of the complete pipeline is more than 100 ms, which is reported in~\cite{BROGGI2012161} as a sufficient planning frequency of an autonomous vehicle. Our st-MLP is a lightweight network that can run in less than 1 ms on our autonomous vehicle, while the complete pipeline including pose estimation takes around 42 ms.

\subsection{Attention-Based Networks}
Attention-based models are common in natural language processing~\cite{vaswani2017attention, devlin2018bert, parikh2016decomposable}, object detection~\cite{ramachandran2019stand, carion2020end} and image classification~\cite{dosovitskiy2020image, wu2020visual}, among others. A class of neural networks that use the attention concept among layer normalization and linear layers are Transformers. Recently, MLP-Mixers have been proposed by~\cite{tolstikhin2021mlp} as an efficient alternative to computationally demanding models like CNNs~\cite{krizhevsky2012imagenet} and self-attention-based Transformers~\cite{vaswani2017attention}, while achieving similar performance on popular image classification benchmarks. The MLP-Mixer architecture is inspired by both CNNs and Transformers since it processes image patches instead of the whole image. It contains channel-mixing blocks for each token (image patch) to capture spatial and per-channel features. Although it has shown competitive performance for image classification, it has not been explored for other data types. Our work is the first to present a model similar to an MLP-Mixer that works on 3D body pose skeletons overtime on the task of gesture recognition in automated driving.

\section{Method}
\label{sec:method}
We focus on skeleton-based gesture recognition where the 3D body pose of an individual serves as input to our approach, as it outperforms image-based methods. We define the spatio-temporal multilayer perceptron (st-MLP) that receives a sequence of 3D body poses as input to perform gesture prediction as a classification task. An overview of our method is shown in Fig.~\ref{fig:overview_stmlp}.

\subsection{Problem Formulation}
\label{subsec:problem_formulation}
Let $\mathcal{D} = \{\mathbf{X}_{i},\mathbf{y}_{i} \}_{i=1}^{N}$ be the train set where each sample consists of $T$ body skeletons $\mathbf{X}_{i}=\{\mathbf{x}_{i,0},\dots,\mathbf{x}_{i,T}\}$ and the associated ground-truth gesture as one-hot vector labels $\mathbf{y}_i \in \{0,1\}^{C}$, such that $\sum_{c=1}^{C}\mathbf{y}_i(c)=1$ for a C-category classification problem. At each time step $t\in T$, the corresponding 3D body skeleton $\mathbf{x}_{i,t} \in \mathbb{R}^{3\times K}$ is represented by $K$ body joints. Based on the train set $\mathcal{D}$, our objective is to learn to predict the gesture category, i.e. $\mathbf{y}_{i}$, for the 3D skeleton input sequence $\mathbf{X}_{i}$. We define this mapping as $\mathbf{y}_{i} = f(\mathbf{X}_{i};\mathbf{\theta})$, where $\mathbf{\theta}$ is a set of learnable parameters. To approximate the mapping function, we propose the st-MLP below.

\begin{figure}[!ht]
\vspace{2mm}
    \centering
    \includegraphics[width=0.45\textwidth]{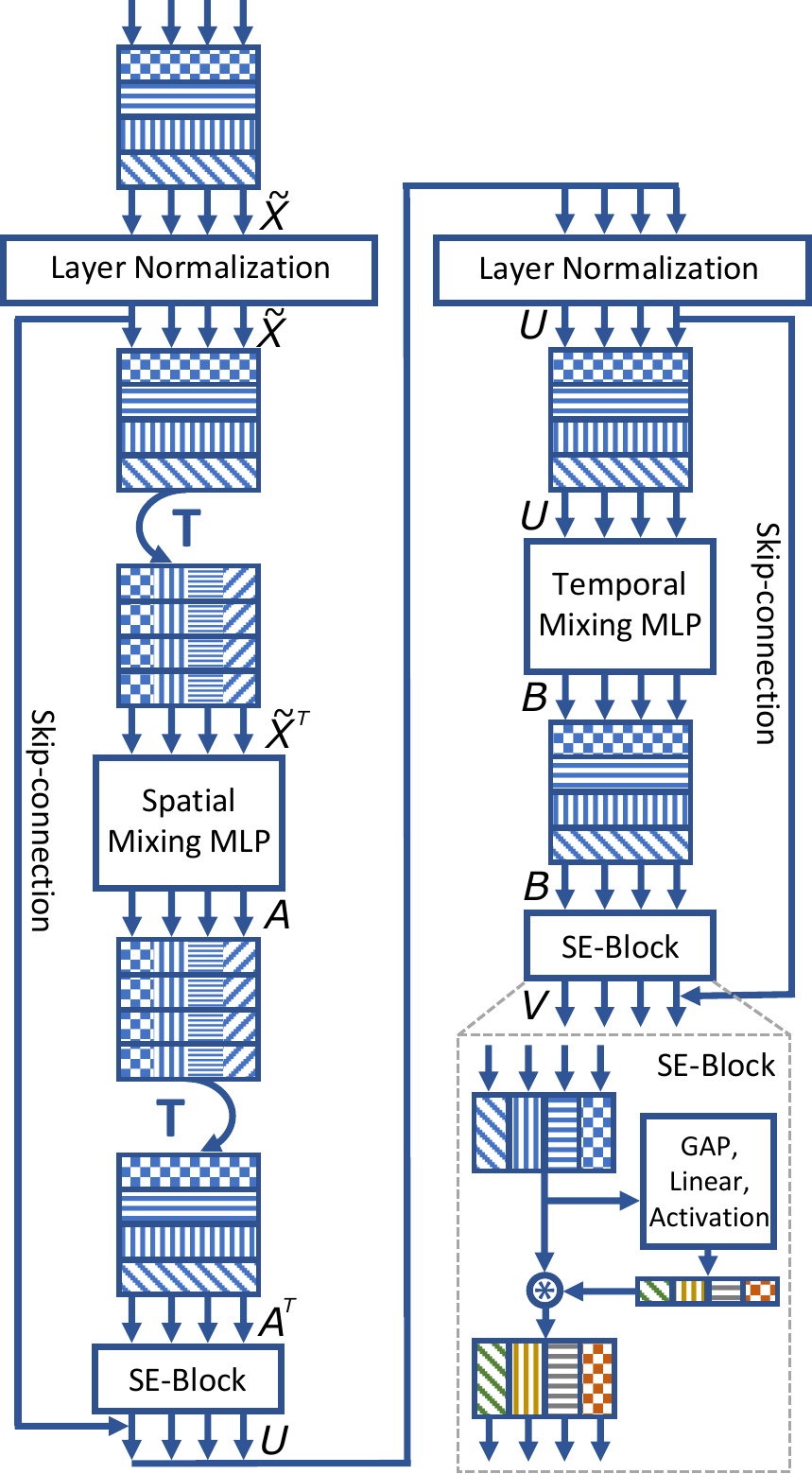}
    \caption{The figure depicts one mixing block, of which the st-MLP can contain several. The mixing block takes an interval of encoded 3D human body skeletons and predicts the presented gesture. The rows of the matrices represent the different time steps, the columns the different joints. In the lower right corner, the architecture of the SE-block is shown. The weight for each channel is calculated with global average pooling (GAP), linear layers, and activation functions.}
    \label{fig:st_mlp}
\end{figure}

\subsection{Spatio-Temporal MLP}
\label{subsec:mlp_gesture_recognition}

Our motivation comes from the concept of MLP-Mixers~\cite{tolstikhin2021mlp}, which is an efficient attention mechanism for image data~\cite{vaswani2017attention, dosovitskiy2020image}. The MLP-Mixer is designed to learn and mix features from images for the classification tasks. To this end, the images are divided into smaller patches of the same size and fed into the mixer model. As the MLP-Mixer performs mixing across all dimensions, the attention is shifted from only considering single patches to also learning the relation between the patches. By mixing a specific dimension, we introduce new features in this dimension. At first, all dimensions are individually mixed and enriched with features. Subsequent mixing operations are then applied to these aggregated features. This results in cross-dimensionally mixed features, similarly to separable convolutions for CNNs, with the added ability of MLPs to aggregate data across the entire dimension instead of a small kernel window. However, unlike images, we deal with a different type of data and therefore present a new formulation to classify a sequence of 3D human skeletons. Additionally, we integrate the Squeeze-and-Excitation (SE) block into our model to re-weight the influence of each time step as mentioned in Sec.~\ref{subsec:se_model}.

In our problem, our model learns to mix features from 3D body skeletons, which are defined in space and time. We define the mixing operation in the time domain as \textit{temporal-mixing} (shown on the right side of Fig.~\ref{fig:st_mlp}) and mixing across the body joints as the \textit{spatial-mixing} operation (shown on the left side of Fig.~\ref{fig:st_mlp}). Given a single input sequence $\mathbf{X}$ divided over $T$ time steps, the skeleton of each time step $\mathbf{x}_{t}\in\mathbb{R}^{K\times 3}$ is flattened into a vector of length $k = 3 * K$. This results in a two-dimensional input $\mathbb{R}^{T\times k}$ containing one temporal dimension and one spatial dimension. Flattening is needed to reduce the two-dimensional joints into one dimension in order to enable spatial mixing. Next, the flattened dimension $k$ is projected into a hidden dimension $S$ using a convolutional layer. The transformed input can be represented as a matrix $\mathbf{\Tilde{X}} \in \mathbb{R}^{T\times S}$. The matrix is passed through each of the $L$ layers of the st-MLP.

Each layer $l \in [1 \hdots L]$ consists of a mixing block containing the \textit{spatial-mixing} unit, followed by the \textit{temporal-mixing} unit. We illustrate a mixing block in Fig.~\ref{fig:st_mlp}. Both types of units contain fully connected layers, i.e. MLPs, and expand the input to a hidden dimension, namely to $D_S$ for the \textit{spatial-mixing} units and $D_T$ for the \textit{temporal-mixing} units. The values of both hidden dimensions can be selected independently of the number of joints and sequence length. In addition, each layer contains skip-connections, non-linear activation functions, and Layer Normalization~\cite{ba2016layer} to obtain meaningful gradients during backpropagation and prevent overfitting. The Layer Normalization is similar to regular Batch Normalization~\cite{pmlr-v37-ioffe15}, however, it normalizes the input over the channel dimension instead of the batch dimension. In our case, this provides normalization over the time dimension. The first unit (\textit{spatial-mixing}) operates on the rows of $\mathbf{\Tilde{X}}$. Therefore, the first unit is applied to the transposed matrix $\mathbf{\Tilde{X}}^T$. The second unit (\textit{temporal-mixing}) operates on the columns of $\mathbf{U}$, which is the output from the spatial-mixing block. The input to the first layer is the information $\mathbf{\tilde{X}}_0$ extracted from 3D skeletons, while $\textbf{V}_0$ is the output. The input $\mathbf{\tilde{X}}_l$ of the following layer $l$ is the output $\textbf{V}_{l-1}$ of the previous layer $l-1$, i.e. $\textbf{V}_{l-1}=\mathbf{\tilde{X}}_{l}$. For the sake of simplicity, we use a single sample to define $\mathbf{U}$ and $\mathbf{V}$ as:
\begin{equation}
\label{eq:spat_mix}
\begin{gathered}
    \textbf{U}_{s,l} =  \mathbf{\tilde{X}}_{s,l} + (\textbf{W}_{2,l}(g(\textbf{W}_{1,l}\, h(\mathbf{\Tilde{X}}^T_{s,l}))))^T, \\\
    s\in [1\dots S],
\end{gathered}
\end{equation}
\begin{equation}
\label{eq:temp_mix}
\begin{gathered}
    \textbf{V}_{j,l} =  \textbf{U}_{j,l} + \textbf{W}_{4,l}(g(\textbf{W}_{3,l}\, h(\mathbf{U}_{j,l}))), \\\
    j\in [1\dots T],
\end{gathered}
\end{equation}

where $g(.)$ is the non-linear (GeLU) activation and $h(.)$ is the layer normalization operator. Finally, the output from the last layer is passed through a global average pooling layer followed by a linear classifier to predict the gesture class $\mathbf{\Hat{y}}$.

\subsection{Squeeze-and-Excitation Block}
\label{subsec:se_model}

The Squeeze-and-Excitation (SE) block~\cite{Hu_2018_CVPR} takes the distinct influence of each channel of a feature tensor into account and assigns a weight to each channel depending on its importance. Since the influence of the time steps can differ between the various gestures, the weights are not fixed and can vary between gestures. This helps the network to take the higher importance of recent time steps compared to earlier time steps into account, which is similar to LSTMs weighting its hidden states and its current input to focus on either history or current values more. An overview of the SE-block is shown in Fig.~\ref{fig:st_mlp}. The SE-block uses global average pooling to condense every feature layer into a single value. In a second step, the compressed feature tensor is processed with a linear layer network which contains two linear layers as well as ReLU and Softmax activation functions. Finally, each feature layer of the original tensor is multiplied with its corresponding weight of the linear layer network output to assign the calculated importance to each channel.

In gesture recognition, each time step has a different influence on the predicted gesture. The most recent time step $T$ of the skeleton sequence $\mathbf{X}$ has a higher influence on the ground truth $\textbf{y}$ than the oldest time step $0$. To take this into account, we use the SE-block to weight the importance of each time step in the model. In Fig.~\ref{fig:st_mlp} we show the integration of the SE-block after the \textit{temporal-mixing} and \textit{spatial-mixing} units into the mixing block but before the addition of the skip connection. Combining Eq.~\ref{eq:spat_mix} and Eq.~\ref{eq:temp_mix} with the SE-block, the importance weighting of the time domain can be included in each layer $l$. For better understanding, we reformulate the new output for a single sample as:

\begin{equation}
\begin{gathered}
    \textbf{A}_{s,l} = \textbf{W}_{2,l}(g(\textbf{W}_{1,l}\, h(\mathbf{\tilde{X}}^T_{s,l}))), \\\
    \textbf{U}_{s,l} = \mathbf{\tilde{X}}_{s,l} + \sigma(\textbf{W}_{2SE} r(\textbf{W}_{1SE}\textbf{A}_{s,l}^T)), \\\
    \: s\in [1\dots S],
\end{gathered}
\end{equation}
\begin{equation}
\begin{gathered}
    \textbf{B}_{j,l} = \textbf{W}_{4,l}(g(\textbf{W}_{3,l}\, h(\mathbf{U}_{j,l}))), \\\
    \textbf{V}_{j,l} =  \textbf{U}_{j,l} + \sigma(\textbf{W}_{2SE} r(\textbf{W}_{1SE}\textbf{B}_{j,l})), \\\
    \: j\in [1\dots T],
\end{gathered}
\end{equation}
where $r(.)$ and $\sigma(.)$ are the ReLU and the Softmax activation functions respectively. $\mathbf{A}_{s,l}$ and $\mathbf{B}_{j,l}$ are the results of the spatial-mixing and temporal-mixing before the addition of the skip connection. The time steps in $\mathbf{A}_{s,l}$ and $\mathbf{B}_{j,l}$ are weighted by the included SE-block. At both locations, we use the same SE-block module resulting in shared SE-block weights $\mathbf{W}_{1SE}$ and $\mathbf{W}_{2SE}$ across both mixer units. This leads to similar time step weights at both SE-block locations, improving model performance.

\subsection{Complete Model}

In Fig.~\ref{fig:st_mlp} we show the structure of one mixing block, where the SE-block is already included right before the skip-connection. The st-MLP consists of several mixing blocks stacked one after another. The number of mixing blocks is defined by the number of layers $L$.

The aim is to minimize the cross-entropy between the predictions $\mathbf{\hat{y}}_{i}$ and ground truth $\mathbf{y}_{i}$. This requires the optimization of the model weights by backpropagation and an optimization algorithm, hence the optimal model weights $\theta^{*}$ should satisfy the following condition:

\begin{equation}
    \theta^{*} = \arg\min_{\theta}  \mathcal{L}(\mathbf{y}_{i}, f(\mathbf{X}_{i};\mathbf{\theta})).
\end{equation}

\section{AUTONOMOUS VEHICLE IMPLEMENTATION}
\label{sec:_auton_car_impl}

We evaluate gesture recognition based on st-MLPs in a real-world scenario on an autonomous vehicle. In the following section, we describe our implementation for the gesture recognition task and the hardware specification of the autonomous vehicle.

\subsection{Gesture Recognition}
\label{subsec:gesture_recognition}
\begin{figure}
    \vspace{2mm}
    \centering
    \includegraphics[width=0.48\textwidth]{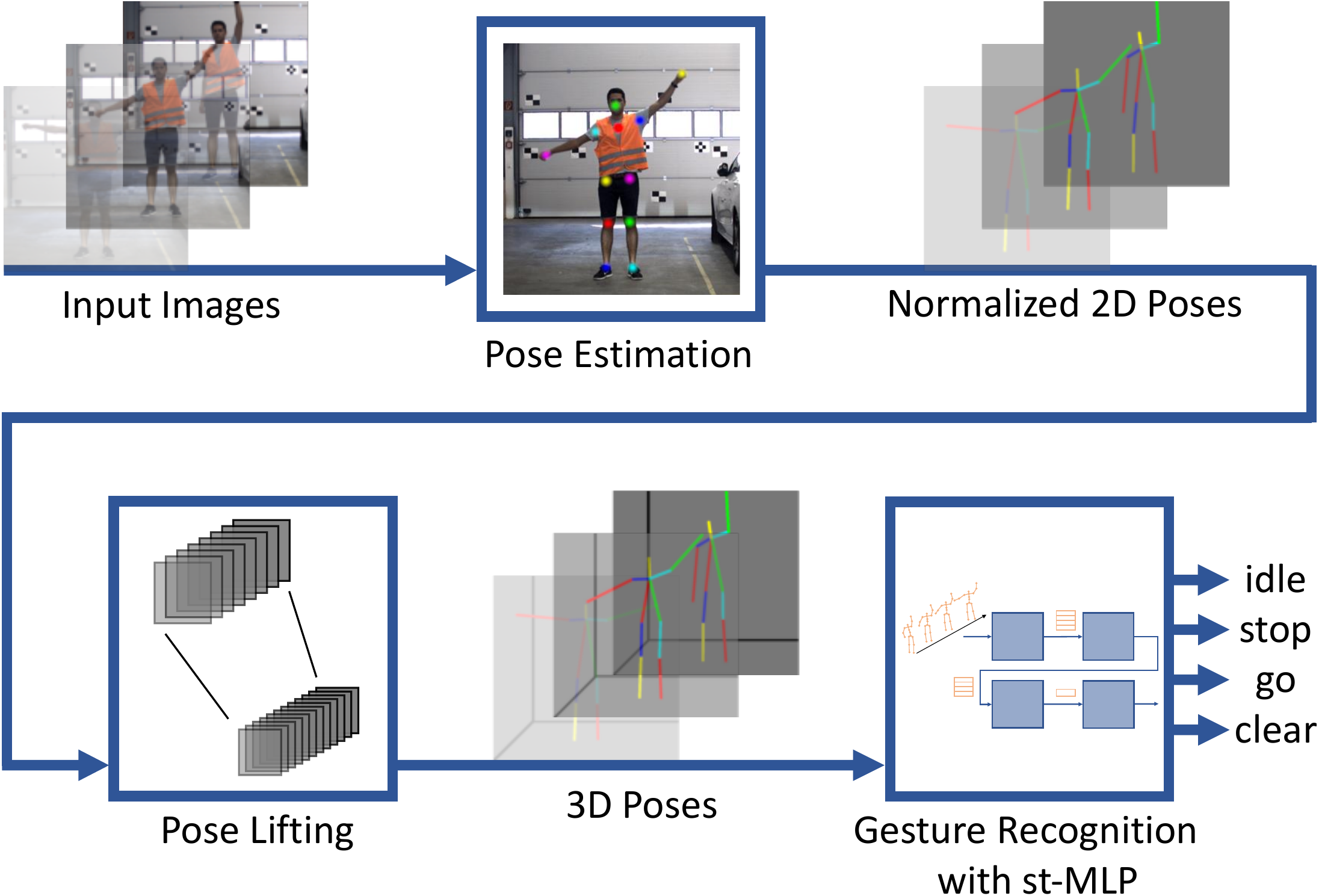}
    \caption{Our proposed gesture recognition pipeline on the autonomous vehicle. Our input consists of sequences of camera images. The images are fed into a pose estimation framework to obtain sequences of 2D poses. These poses are then lifted with another deep neural network to 3D poses and then processed with our st-MLP to predict the gesture.}
    \label{fig:real_world_demo}
\end{figure}

The st-MLP processes 3D human body skeletons to predict the presented gesture. However, there are no sensors available for autonomous vehicles that directly return human body skeletons. Therefore, we use an RGB camera mounted on a vehicle to perform gesture recognition. An overview of the full approach is shown in Fig.~\ref{fig:real_world_demo}. To extract a 2D human body skeleton from the camera image, a recent deep learning approach for human pose estimation called OpenPifPaf~\cite{Kreiss_2019_CVPR} is used. The pose extraction step provides 2D body pose estimates which are lifted using the VideoPose3D-CNN~\cite{pavllo:videopose3d:2019} to obtain 3D body pose estimates. The lifting procedure helps to resolve ambiguities coming from certain 2D poses, as well as the orientation variance of specific gestures. We accumulate and stack the 3D human body poses $\mathbf{x}$ of the last $T$ time steps to get a sequence of poses $\mathbf{x}_{0}, \dots, \mathbf{x}_{T}$. This sequence is fed into the st-MLP to predict the gesture $\mathbf{\Hat{Y}}_{T}$ performed at time step $T$.

\subsection{Specification of the Autonomous Vehicle}
\label{subsec:car_specs}
The test vehicle is equipped with an Intel Xeon E5-2640 CPU and NVIDIA 2080Ti GPU, running Ubuntu 18.04 in a docker container. To interact in urban environments the autonomous vehicle must be able to react in real-time to changes in the environment. As most of the time is used for neural network inference, we rely on the NVIDIA TensorRT framework~\cite{tensorrt} to speed up execution. The main advantage of TensorRT is the provided execution optimization which improves the memory latency and allocation. Furthermore, it is used to convert the model parameters from 32-bit to 16-bit floating-point precision to speed up calculations. By using our st-MLP with fixed input sizes, we achieve quicker and more stable execution times using TensorRT compared to standard LSTMs, which is highly important in a real-time environment.

\section{EXPERIMENTS}
\label{sec:experiments}
We first perform evaluations on two standard benchmarks and then on real-world settings. Furthermore, we introduce the evaluation datasets, provide the training details of our st-MLP, and show the effect of different components of our approach with ablation studies. 

\subsection{Datasets}
\label{subsec:datasets}
For the evaluation, we use two gesture recognition datasets for autonomous driving. The first dataset is the TCG dataset~\cite{wiederer2020traffic}. The TCG dataset contains the standard European traffic control gestures, namely \textit{Stop}, \textit{Go} and \textit{Clear}. Additionally, a fourth class named \textit{Idle} is added. The Idle gesture resembles the situation where none of the previous three classes is present. The dataset contains gestures from different scenarios at 4-way-intersections and T-intersections as well as different camera views of the traffic controller. These different viewpoints affect the label, e.g. the label of the stop gesture differs if the traffic controller is seen from the front or the side. The gestures are performed by 5 different subjects which are represented by 17 human body joints stacked into a time sequence. The evaluation is performed on the test set following the TCG protocol in a cross-view and cross-subject manner based on the accuracy, Jaccard index, and F1-score metrics.

The second dataset in the evaluation is the Drive\&Act dataset~\cite{drive_and_act_2019_iccv}. This dataset is recorded inside of a car and contains gestures and actions that can be performed while driving, e.g. \textit{eating/drinking}, or actions becoming possible with an increasing amount of automated driving like \textit{reading a newspaper}. The gestures in the dataset are executed twice by 15 different subjects. The subjects are represented by a skeleton consisting of 13 joints (4 fewer joints compared to TCG), as the lower part of the body has no influence on the performed gesture, and leg detection with cameras is impossible due to occluded perspectives. For the evaluation, we follow the protocol for coarse classes on the validation and test set as defined in the Drive\&Act dataset and use three different splits for each of the training, validation, and test sets. The mean per-class accuracy is the evaluation metric. 

The main differences between these datasets are the different number of classes, base skeletons, and action locations, as the actions contained in Drive\&Act are performed inside of a car while the gestures from the TCG dataset are outdoors. Furthermore, we classify the entire sequence with one label for Drive\&Act, while for TCG an action is predicted for each time step.

In addition to the comparison of st-MLP with other gesture recognition approaches, we show the real-time capability of our st-MLP on the autonomous vehicle described in Sec.~\ref{subsec:car_specs}. Therefore, we rely on a labeled sequence recorded with our autonomous vehicle. We annotate this sequence with the labels of the TCG dataset. Note that we report the mean inference time and gesture recognition performance over 167 frames at 15 fps based on a $1024 \times 768$ image resolution.

\subsection{Spatio-Temporal MLP Implementation}
\label{subsec:st_mlp}
The st-MLP is implemented using PyTorch~\cite{pytorch}. At the input, a 1D convolution of kernel size $1 \times k$ is applied. Further processing is done with linear layers, GeLU activation functions, transpose operations, and the SE-block. The output of the last linear layer goes through the softmax activation function for gesture classification.

The TCG dataset is trained for 70 epochs with a balanced batch (i.e.~equal class probability for each sample) of size 1024. For the optimization, the Ranger optimizer~\cite{Ranger} with a flat and cosine annealing learning rate scheduler is used. After the first 50 epochs, the learning rate of 0.001 is reduced to 0.0001 with the cosine annealing function. As for the internal hyperparameters of our st-MLP model, the number of layers $L$ is set to 4, the hidden dimension $S$ to 512, the sequence length $T$ to 24, the hidden dimension for the spatial-mixing $D_S$ to 32, and the hidden dimension for the temporal-mixing $D_T$ to 256. 

The optimization on the Act\&Drive dataset is done with different hyperparameters due to varying settings, e.g. the number of joints and number of classes. The training takes place for 80 epochs with a balanced batch of size 2048. We use the Adam optimizer~\cite{kingma2014adam} with a learning rate of 0.001 and a cosine annealing learning rate scheduler to reduce the learning rate by a factor of 0.1 at the end of training. We use the same model hyperparameters for the st-MLP as in the TCG dataset, only changing the number of layers $L$ to 2, the sequence length $T$ to 90, and the hidden dimension for the spatial-mixing $D_S$ to 64. Additionally, we convert the world coordinates of the dataset into camera coordinates.

The aforementioned hyperparameters controlling architecture depth/hidden dimensions, learning rate, and training epochs are optimized using grid search for each dataset separately. Furthermore, for each method the best performing parameters have been selected for our experiments.

\subsection{Evaluation of the Spatio-Temporal MLP}
\label{subsec:eval_mlp}

We present separately below the evaluation on the TCG and Drive\&Act datasets.

\paragraph{TCG Dataset}
\label{para:tcg_dataset_eval}
We compare our st-MLP with other models~\cite{wiederer2020traffic} for gesture recognition in the cross-subject and the cross-view protocol on the TCG dataset. The results are shown in Tab.~\ref{tab:cross_subj_tcg} for the cross-subject evaluation and in Tab.~\ref{tab:cross_view_tcg} for the cross-view evaluation. In both tables, we provide the mean and standard deviation over 3 runs, similar to~\cite{wiederer2020traffic}. The accuracy of the st-MLP in both evaluation protocols is slightly below the best accuracy of the other models. In contrast, the Jaccard index and the F1-score of the st-MLP are better than the other models.

\begin{table}
\renewcommand{\arraystretch}{1.1}
\vspace{2mm}
\centering
\caption{Cross-Subject evaluation of our spatio-temporal MLP (st-MLP) on the test set of the TCG dataset. Results of other methods used from~\cite{wiederer2020traffic}.}
\label{tab:cross_subj_tcg}
    \begin{tabular}{p{2cm}|C{1.5cm}C{1.5cm}C{1.5cm}}
    \toprule
    \multicolumn{1}{c|}{\multirow{3}{*}{\textbf{Method}}} & \multicolumn{3}{c}{\textbf{Metric}} \\
    \multicolumn{1}{c|}{} & Accuracy & Jaccard Index & F1-Score \\ 
    \midrule
    GRU & 84.44(\textit{\scriptsize$\pm$\scriptsize 2.0}) & 58.16(\textit{\scriptsize$\pm$\scriptsize 4.2}) & 70.45(\textit{\scriptsize$\pm$\scriptsize 3.1}) \\
    Att-LSTM & 85.67(\textit{\scriptsize$\pm$\scriptsize 2.1}) & 50.70(\textit{\scriptsize$\pm$\scriptsize 9.9}) & 61.87(\textit{\scriptsize$\pm$\scriptsize 10.6}) \\
    Bi-GRU & 86.80(\textit{\scriptsize$\pm$\scriptsize 1.6}) & 57.25(\textit{\scriptsize$\pm$\scriptsize 7.4}) & 68.95(\textit{\scriptsize$\pm$\scriptsize 6.4}) \\
    Bi-LSTM & \textbf{87.24}(\textit{\scriptsize$\pm$\scriptsize 1.8}) & 67.00(\textit{\scriptsize$\pm$\scriptsize 2.1}) & 78.48(\textit{\scriptsize$\pm$\scriptsize 1.8}) \\
    st-MLP (ours) & 85.99(\textit{\scriptsize$\pm$\scriptsize 0.11}) & \textbf{67.88}(\textit{\scriptsize$\pm$\scriptsize 0.06}) & \textbf{80.05}(\textit{\scriptsize$\pm$\scriptsize 0.03}) \\ 
    \bottomrule
    \end{tabular}
\end{table}

\begin{table}
\renewcommand{\arraystretch}{1.1}
\centering
\caption{Cross-View evaluation of our spatio-temporal MLP (st-MLP) on the test set of the TCG dataset. Results of other methods used from~\cite{wiederer2020traffic}.}
\label{tab:cross_view_tcg}
    \begin{tabular}{p{2cm}|C{1.5cm}C{1.5cm}C{1.5cm}}
    \toprule
    \multicolumn{1}{c|}{\multirow{3}{*}{\textbf{Method}}} & \multicolumn{3}{c}{\textbf{Metric}} \\
    \multicolumn{1}{c|}{} & Accuracy & Jaccard Index & F1-Score \\ 
    \midrule
    GRU & 83.47(\textit{\scriptsize$\pm$\scriptsize 1.4}) & 56.25(\textit{\scriptsize$\pm$\scriptsize 7.6}) & 68.59(\textit{\scriptsize$\pm$\scriptsize 7.4}) \\
    Att-LSTM & 85.30(\textit{\scriptsize$\pm$\scriptsize 1.1}) & 59.87(\textit{\scriptsize$\pm$\scriptsize 12.7}) & 71.20(\textit{\scriptsize$\pm$\scriptsize 12.3}) \\
    Bi-GRU & \textbf{87.37}(\textit{\scriptsize$\pm$\scriptsize 0.3}) & 55.55(\textit{\scriptsize$\pm$\scriptsize 2.8}) & 67.68(\textit{\scriptsize$\pm$\scriptsize 2.2}) \\
    Bi-LSTM & 86.66(\textit{\scriptsize$\pm$\scriptsize 1.2}) & 65.95(\textit{\scriptsize$\pm$\scriptsize 4.7}) & 77.14(\textit{\scriptsize$\pm$\scriptsize 4.3}) \\
    st-MLP (ours) & 86.90(\textit{\scriptsize$\pm$\scriptsize 0.08}) & \textbf{70.83}(\textit{\scriptsize$\pm$\scriptsize 0.28}) & \textbf{82.48}(\textit{\scriptsize$\pm$\scriptsize 0.21}) \\ 
    \bottomrule
    \end{tabular}
\end{table}

\begin{figure}
    \centering
    \subfloat[Cross-subject; Bi-LSTM\label{fig:xs_bilstm}]{%
      \includegraphics[width=0.2\textwidth]{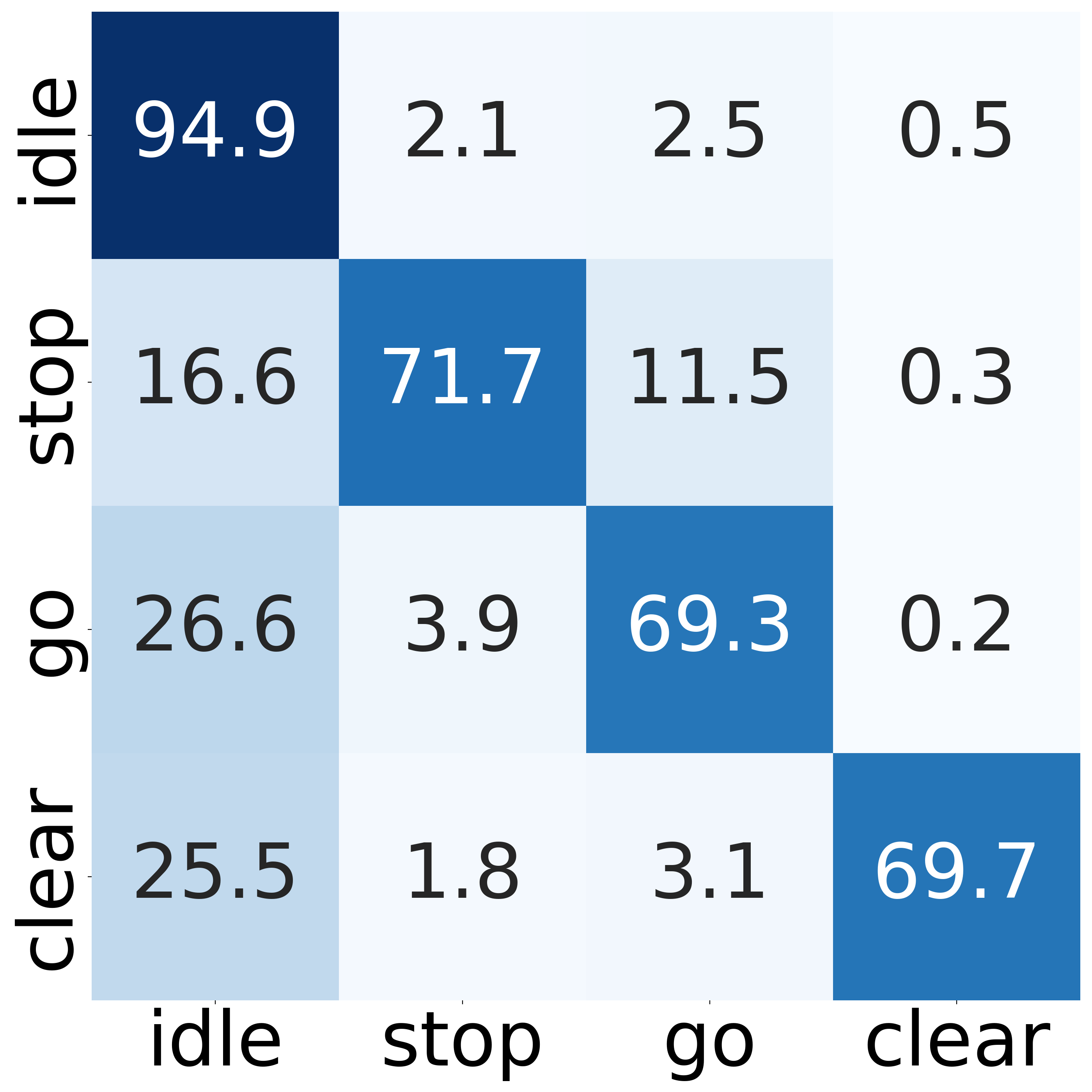}
    }
    \hspace{2mm}
    \subfloat[Cross-view; Bi-LSTM\label{fig:xv_bilstm}]{%
      \includegraphics[width=0.2\textwidth]{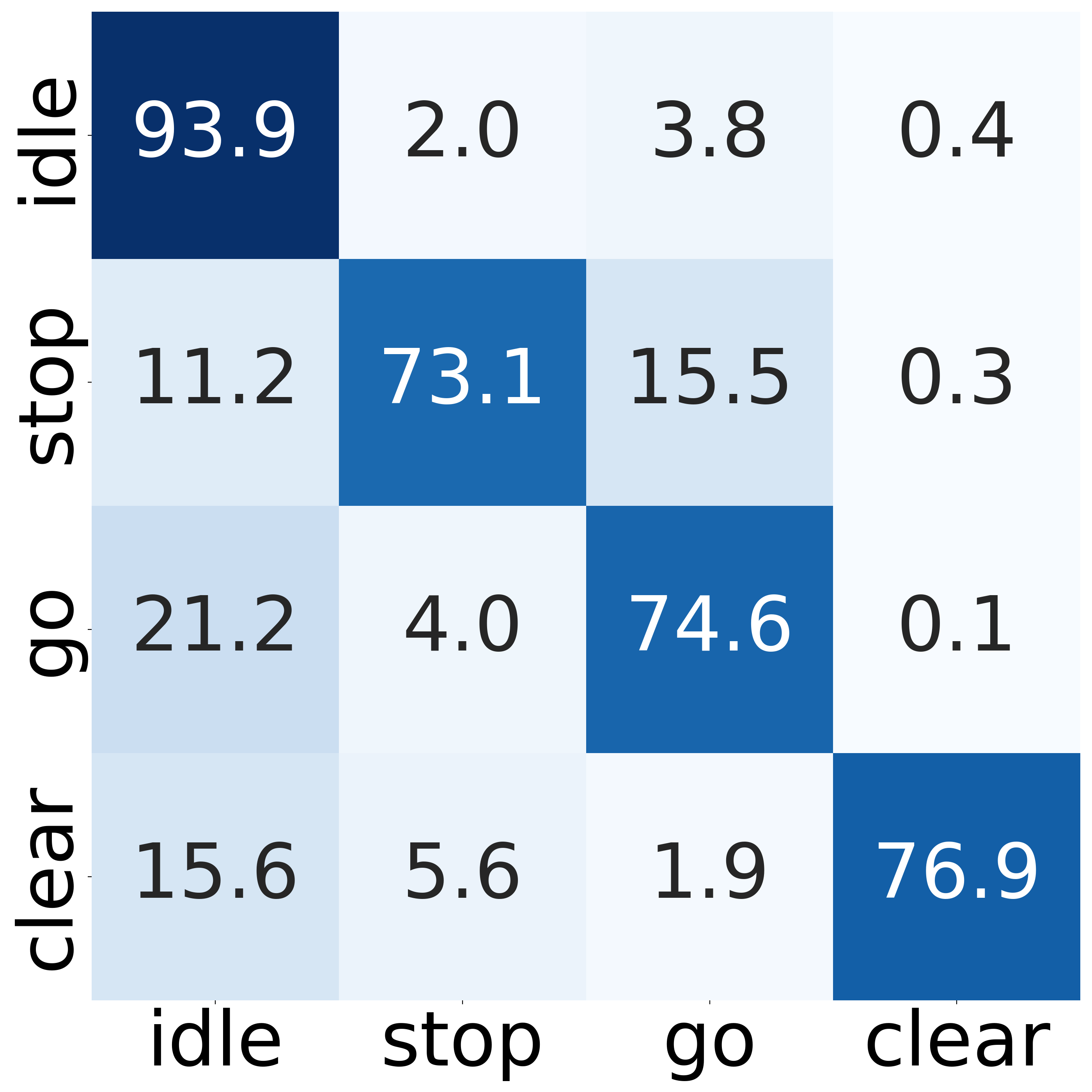}
    }
    \\
    \subfloat[Cross-subject; st-MLP\label{fig:xs_stmlp}]{%
      \includegraphics[width=0.2\textwidth]{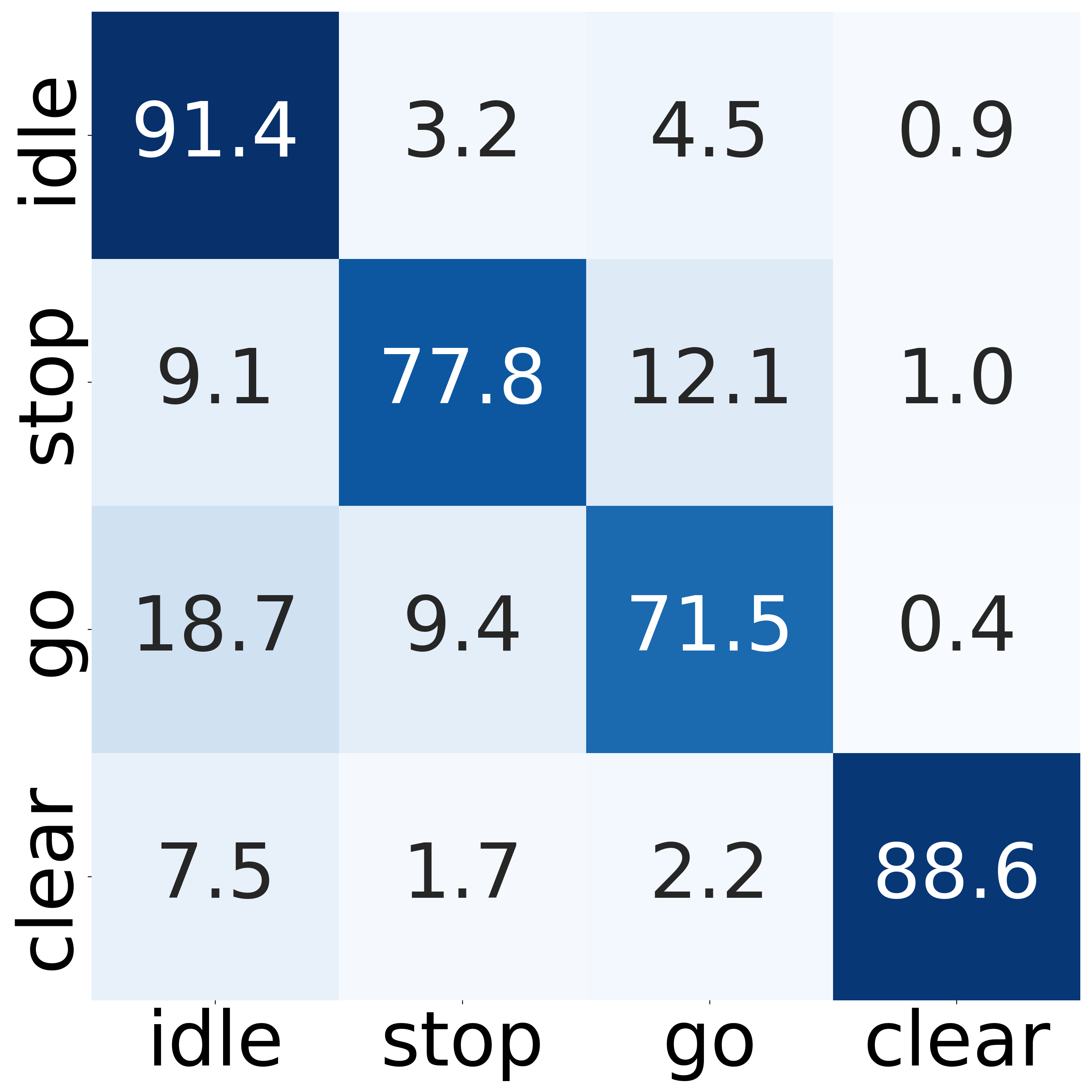}
    }
    \hspace{2mm}
    \subfloat[Cross-view; st-MLP\label{fig:xv_stmlp}]{%
      \includegraphics[width=0.2\textwidth]{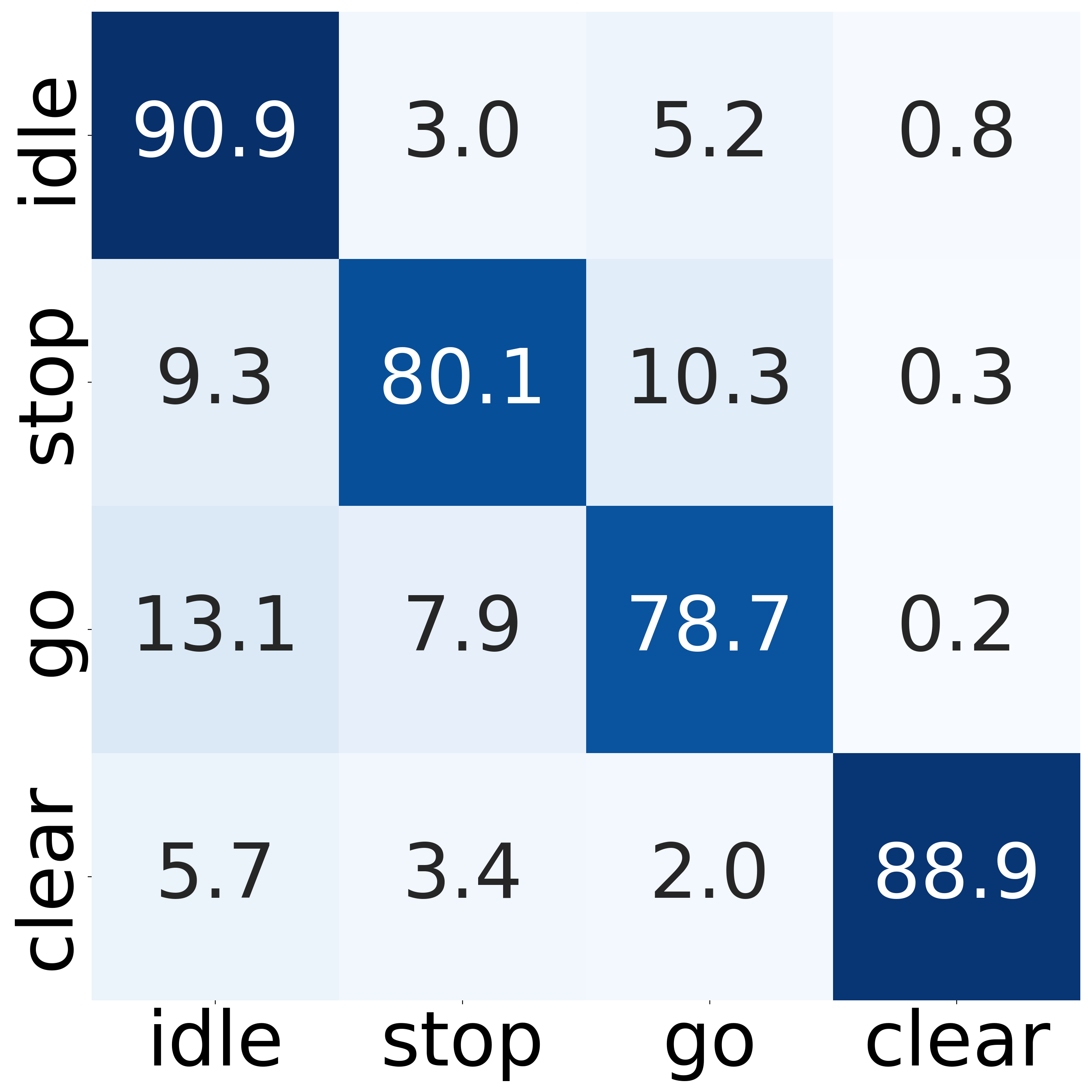}
    }
    \caption{Confusion matrices of the Bi-LSTM used from~\cite{wiederer2020traffic} and our spatio-temporal MLP (st-MLP) for the cross-subject and cross-view task of the TCG dataset.}
    \label{fig:confusion_matrices_tcg}
\end{figure}

To understand the drop in the accuracy and the increase in the other metrics, we plot the confusion matrices of our st-MLP and the confusion matrix of the best working model from~\cite{wiederer2020traffic}, namely the Bi-LSTM in Fig~\ref{fig:confusion_matrices_tcg}. Compared to the Bi-LSTM, shown in Fig.~\ref{fig:xs_bilstm} and Fig.~\ref{fig:xv_bilstm}, the st-MLP shown in Fig.~\ref{fig:xs_stmlp} and Fig.~\ref{fig:xv_stmlp}, has a lower accuracy in the Idle class but improves the detection performance on the other classes. Due to the high imbalance of the dataset (many samples belong to the idle class, while fewer samples belong to the other classes), the accuracy drops with lower accuracy in the idle class. This is caused by the weighted mean calculation of the accuracy metric. To this end, relying on accuracy alone is insufficient to assess the model performance on TCG, hence, F1-score and Jaccard index are additionally calculated per label and then summarized by an unweighted mean. This leads to higher detection performance for under-represented classes and therefore better results for the F1-score and Jaccard index. Furthermore, our approach has a smaller standard deviation during three different evaluation runs compared to the baseline as shown in Tab.~\ref{tab:cross_subj_tcg} and Tab.~\ref{tab:cross_view_tcg}. This indicates that our approach delivers more consistent results.

\paragraph{Drive\&Act Dataset}
\label{para:aad_dataset_eval}
To examine the generalization abilities of our st-MLP, we evaluate on a different autonomous driving dataset, namely the Drive\&Act dataset~\cite{drive_and_act_2019_iccv}. In Tab.~\ref{tab:eval_aad}, we compare our results on the evaluation protocol of the Drive\&Act dataset with the results of the paper's temporal models. In the columns \textit{Validation} and \textit{Test}, the best mean per-class accuracy results on the validation and test sets are shown. We only compare with the Pose method of the Drive\&Act dataset, as only in this method the same input data is used. In the Two-Stream approach, additional spatial dependencies between the joints are employed. Comparing the st-MLP with the Pose approach, we gain 1.65\% accuracy. Furthermore, we reach similar results to the Two-Stream method by only using the same data as the Pose method.

\begin{table}
\renewcommand{\arraystretch}{1.1}
\centering
\vspace{2mm}
\caption{Evaluation of the spatio-temporal MLP (st-MLP) on the Drive\&Act dataset. Results of the other methods were obtained from \cite{drive_and_act_2019_iccv}.}
\label{tab:eval_aad}
    \begin{tabular}{p{2cm}|C{2cm}C{2cm}}
        \toprule
        \textbf{Method} &  \textbf{Validation} & \textbf{Test}\\
        \midrule
        Pose~\cite{drive_and_act_2019_iccv} & 37.18 & 32.96 \\
        Two-Stream~\cite{drive_and_act_2019_iccv} & 39.37 & \textbf{34.81} \\
        st-MLP (ours) & \textbf{40.56} & 34.61 \\ 
        \bottomrule
    \end{tabular}
\end{table}

\subsection{Ablation Studies}
\label{subsec:ablation_studies}
Each component of the st-MLP has a different effect on the model performance. We show the effect of \textit{spatial-mixing} and \textit{temporal-mixing} as well as the influence of the SE-block.

\paragraph{Influence of Temporal-Mixing and Spatial-Mixing}
In each layer of the st-MLP, we perform \textit{spatial-mixing} and \textit{temporal-mixing}. We determine the influence of each mixing type on the overall performance by training models with only one mixing type, either \textit{spatial-mixing} or \textit{temporal-mixing}. Furthermore, we evaluate the combination of \textit{temporal-mixing} and \textit{spatial-mixing} in two separate streams, which we call \textit{two-stream} model.

The performance of the single mixing type models is compared with the performance of our st-MLP on the Drive\&Act dataset. The st-MLP has a mean per-class accuracy of 34.61\%. This is around 2.5\% better than the performance of 32.08\% of the model using only \textit{temporal-mixing}, while the model using only \textit{spatial-mixing} achieved 30.98\%. This shows that the combination of both mixing types improves the model. Furthermore, we implement a two-stream model with the \textit{temporal-mixing} stream and \textit{spatial-mixing} stream going through separate neural networks. Before classification, both streams are combined using global average pooling. When comparing the two-stream model with our st-MLP, i.e.~single stream, we find that the alternating st-MLP with a mean per-class accuracy of 34.61\% is superior to the performance of the two-stream model with only 30.09\% accuracy.

\paragraph{Influence of the SE-Block}
\label{para:eval_se_block}

We use the SE-block in our st-MLP to weight different time steps by their importance as described in Sec.~\ref{subsec:se_model}. The st-MLP shares the SE-block module between the \textit{spatial-mixing} and the \textit{temporal-mixing} to reach mean per-class accuracy of 34.61\% on the Drive\&Act dataset. Without a SE-block using the standard MLP-Mixer architecture the st-MLP reaches 0.79\% less mean per-class accuracy. When using two separate SE-blocks behind the \textit{spatial-mixing} and \textit{temporal-mixing}, we are still 0.46\% mean per-class accuracy behind the performance of the st-MLP with shared SE-blocks.

\subsection{Evaluation of the Real World Scenario}
\label{subsec:inference}

In Sec.~\ref{subsec:gesture_recognition}, we present a method to extract 3D human body skeletons from a stream of camera images and perform gesture recognition. To improve real-time execution and achieve constant run-time on our autonomous vehicle, we utilize TensorRT~\cite{tensorrt} for execution. In Tab.~\ref{tab:inference_time}, we present the inference time in milliseconds (ms) for each element of the method. All results are based on the hardware specified in Sec.\ref{subsec:car_specs} and the evaluation sequence from Sec.~\ref{subsec:datasets}. As it can be observed in Tab.~\ref{tab:inference_time}, the inference time of our implementation on the autonomous vehicle is about 42 ms. This means that the prediction of the presented gestures can be easily performed inside of one planning cycle of the vehicle that takes 100 ms according to~\cite{BROGGI2012161}. Therefore, the prediction arrives in time for being considered in the current planning cycle. Without using TensorRT, the inference constantly exceeded 100 ms, while achieving the same accuracy of 83.23\% on our test sequence. Moreover, TensorRT does not impact the model output. One can also see, that the st-MLP part of the pipeline is extremely fast on its own, achieving sub-1ms inference and thus making it suitable for timing-critical environments.

\begin{table}
\renewcommand{\arraystretch}{1.1}
\vspace{2mm}
\centering
\caption{We measure inference times of our optimized approach on the autonomous vehicle described in Sec.~\ref{subsec:car_specs}. The execution time of each part of our recognition system is reported.}
\label{tab:inference_time}
    \begin{tabular}{c|c}
    \toprule
    \textbf{Part} & \textbf{Time (ms)} \\
    \midrule
    2D Pose & 36.06 \\
    3D Pose Lifting & 5.13 \\
    Gesture Recognition & 0.66 \\
    \midrule
    Total & 41.85 \\
    \bottomrule
    \end{tabular}
\end{table}

\section{CONCLUSIONS}

We presented the spatio-temporal multilayer perceptron for skeleton-based gesture recognition in the context of autonomous vehicles. We introduced the \textit{spatial-mixing} and \textit{temporal-mixing} of the 3D body pose overtime to train a gesture classification model. In our evaluations, we reached state-of-the-art performance on the TCG and Drive\&Act datasets. Finally, we deployed our model on our autonomous vehicle to show its real-time capability and stable execution, which are both important features for operation on an autonomous vehicle.








\bibliographystyle{ieeetran}
\bibliography{references}

\end{document}